\pdfoutput=1
\documentclass[10pt, logo, copyright]{nvidiatechreport}

\usepackage[authoryear,round]{natbib}

\definecolor{nvidiagreen}{RGB}{118,185,0}
\hypersetup{allcolors=nvidiagreen}
\makeatletter
\renewcommand{\abscontent}{%
    \par\vskip6pt
    {\noindent\bfseries\fontsize{12}{14}\selectfont Abstract\par}
    \vskip6pt
    \noindent
    \parbox{\dimexpr\linewidth}{\absfont \theabstract}
    \@ifundefined{@keywords}{}{
        \vskip1em \noindent \keywordsfont  Keywords: \@keywords}
}
\makeatother

\definecolor{batrowgreen}{RGB}{211,226,184}
\definecolor{deltagray}{RGB}{224,224,224}
\definecolor{deltaPositive}{RGB}{0,110,60}
\definecolor{deltaNegative}{RGB}{190,30,45}

\newcommand{\BaTAgentRuntime}{OpenHands}
\newcommand{\BaTFourBModel}{BaT-4B}
\newcommand{\BaTNineBModel}{BaT-9B}
\newcommand{\BaTFourBAgent}{BaT-4B Agent}
\newcommand{\BaTNineBAgent}{BaT-9B Agent}

\newcommand{\BaTEvaluationTracks}{7}
\newcommand{\BaTEvaluationRepeats}{10}
\newcommand{\BaTEvaluationCells}{70}
\newcommand{\BaTAverageTurns}{33}

\newcommand{\BaTAgentNineOverall}{79.6}
\newcommand{\OpusAgentOverall}{77.5}
\newcommand{\BaTOpusMargin}{2.1}
\newcommand{\BaTAgentNineABRA}{70.6}
\newcommand{\ABRALeader}{79.9}
\newcommand{\BaTABRAGap}{9.3}
\newcommand{\BaTAgentNineMedText}{50.2}
\newcommand{\MedTextLeader}{65.0}
\newcommand{\BaTMedTextGap}{14.8}
\newcommand{\BaTAgentFourAutoMed}{45.8}
\newcommand{\BaTAgentFourABRA}{39.1}
\newcommand{\BaTAgentFourMedText}{19.2}

\newcommand{\BaTStateBankRows}{20{,}299}
\newcommand{\BaTCleanEERows}{1{,}008}
\newcommand{\BaTFourBSFTRows}{4{,}608}
\newcommand{\BaTNineBSFTRows}{4{,}608}
\newcommand{\BaTMatchedAblationRows}{275}
\newcommand{\BaTRolloutsPerPrompt}{4}

\newtheorem{proposition}{Proposition}

\title{BaT: Towards Self-Evolving Medical Research Agent with Stage Rubrics}

\author[2]{Junqi Liu}
\author[1]{Yufan He}
\author[1]{Yexiao He}
\author[1]{Pengfei Guo}
\author[1]{Dong Yang}
\author[1]{Andriy Myronenko}
\author[1]{Can Zhao}
\author[1]{Hanrong Ye}
\author[2]{Tianhao Qi}
\author[2]{Yuyin Zhou}
\author[1]{Daguang Xu}
\author[1]{Yucheng Tang}

\medskip

\affil[1]{NVIDIA}
\affil[2]{University of California, Santa Cruz}

\begin{abstract}
Long-horizon agents are beginning to automate complete workflows that produce code, reports, and research artifacts.
Medical imaging workflows are multi-stage and data-sensitive, while expert trajectories remain scarce and difficult to share.
Structured benchmarks can localize failures through stage-level rubrics, but standard post-training discards these diagnostics before the next training round.
We present Benchmark-as-Teacher (BaT), a recursive self-improvement system for agent post-training.
BaT contains two linked components: the asynchronous \textbf{Stage Bank} data pipeline and \textbf{BiCuRL} (\textbf{Bi}level \textbf{Cu}rriculum \textbf{R}einforcement \textbf{L}earning), its self-improving post-training method.
Stage Bank synthesizes content-isolated training states outside the policy-update loop.
BiCuRL uses a fixed held-out evaluation to select the next stage curriculum, verifies rollouts with task rubrics, updates the policy with GRPO, and returns the candidate checkpoint to evaluation.
On AutoMedBench-Lite, \BaTFourBModel{} and \BaTNineBModel{} more than double the Overall scores of their Qwen Instruct baselines.
\BaTNineBAgent{} reaches \BaTAgentNineOverall{} Overall, exceeding Claude Opus 4.6 with Claude Code at \OpusAgentOverall{}.
We release the code at \url{https://github.com/AutoMedBench/Benchmark-as-Teacher}.
\end{abstract}

\begin{document}

\maketitle
\vspace{-10pt}
\begin{figure}[!h]
    \centering
    \includegraphics[width=\linewidth]{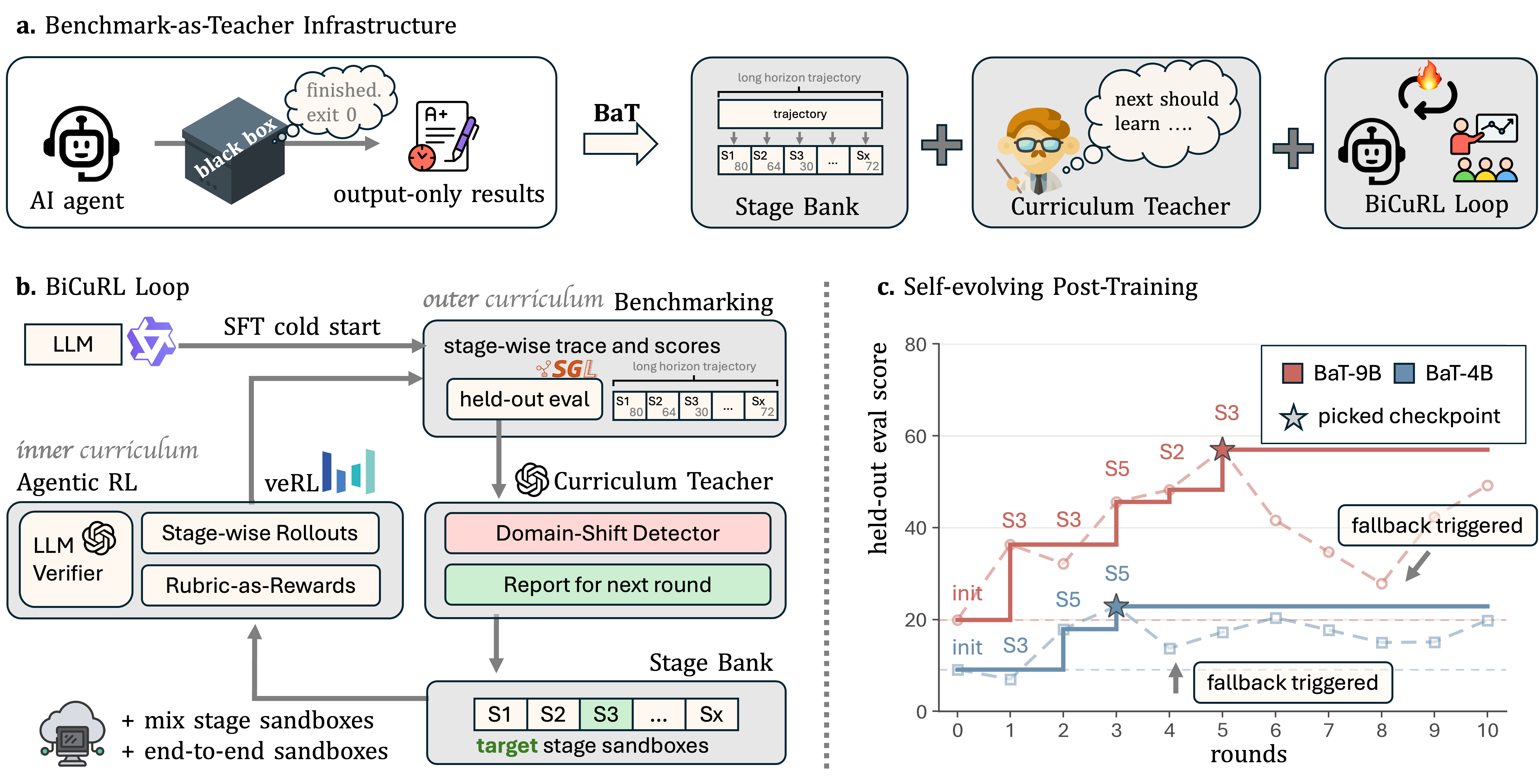}
    \caption{\textbf{Benchmark-as-Teacher turns an output-only benchmark into post-training infrastructure.}
    \textbf{(a) Top: conceptual overview.}
    BaT turns the usual output-only benchmark into training infrastructure with the Stage Bank data-synthesis pipeline, Curriculum Teacher, and BiCuRL post-training method.
    \textbf{(b) Lower left: BiCuRL loop.}
    The system forms a closed loop with four operational phases.
    First, after an SFT cold start, benchmarking splits into stage-wise scores.
    Second, the aggregate scores feed the Curriculum Teacher, shown as the Domain-Shift Detector and Report for next round.
    Third, the environment sandbox pool selects targeted-stage (S-target) sandboxes together with mix-stage (S-mix) and end-to-end (E2E) sandboxes.
    Fourth, the agent executes rollouts in these sandboxes, an LLM verifier scores them with rubric-as-rewards~\citep{gunjal2026rubrics}, and GRPO~\citep{shao2024deepseekmath} updates the policy.
    The updated checkpoint loops back to benchmarking and restarts the cycle.
    \textbf{(c) Lower right: training rounds.}
    Dashed curves show each round's evaluation, solid step lines track the best checkpoint so far, stars mark the auto-selected best checkpoints.
    }
    \label{fig:method-overview}
\end{figure}

\section{Introduction}
\label{sec:introduction}

Long-horizon medical agents must plan, configure tools, validate data, run inference, and submit a checked artifact.
A mistake in one stage can invalidate the work that follows, while expert trajectories remain scarce and difficult to share.
Medical-agent benchmarks increasingly expose the structure of these workflows.
MedAgentBench evaluates 300 physician-authored EHR tasks in a FHIR-compliant environment, HealthAgentBench covers 54 healthcare tasks across seven categories, and AutoMedBench scores planning, setup, validation, inference, and submission~\citep{jiang2025medagentbench,liu2026healthagentbench,liu2026automedbench}.
These benchmarks identify the stage where an agent fails.

Standard post-training discards most of that signal.
Group-relative policy optimization (GRPO) commonly assigns one outcome reward to a complete trajectory~\citep{shao2024deepseekmath}.
A single AutoMedBench-Lite run averages \BaTAverageTurns{} interaction turns, so one score cannot identify which stage needs more practice.
Iterative methods such as Self-Rewarding Language Models and SPIN improve a model over repeated updates, but their training schedules omit held-out stage diagnostics~\citep{yuan2024selfrewarding,chen2024spin}.
This paper asks one question: \emph{can a structured benchmark teach an agent while its task content stays outside training?}

Benchmark-as-Teacher (BaT) turns the benchmark signal into a recursive post-training cycle.
BaT contains an asynchronous data pipeline, called Stage Bank, and a self-improving post-training method, called Bilevel Curriculum Reinforcement Learning (BiCuRL).
Stage Bank synthesizes fictional tasks, reconstructs executable stage states, and applies leakage checks outside the policy-update loop.
It exposes three content-isolated training pools: S-target for the selected weak stage, S-mix for the remaining stages, and an end-to-end (E2E) surrogate for the complete workflow.

BiCuRL connects evaluation, training, and re-evaluation.
Its outer loop reads stage scores from a fixed held-out evaluation, selects the next target stage, and retains or rejects candidate checkpoints.
Its inner loop samples Stage Bank states, scores new rollouts with rubric items and artifact evidence, and updates the policy with GRPO.
Only aggregate scores cross the evaluation boundary; task IDs, answers, paths, reports, and traces remain held out.
Figure~\ref{fig:method-overview} summarizes this RSI cycle.

A trained policy becomes a BaT Agent when paired with a fixed execution environment that includes public stage skills.
We keep this engineering layer fixed across BaT Agent comparisons so BiCuRL changes only the policy.

\begin{figure}[t]
    \centering
    \includegraphics[width=\columnwidth]{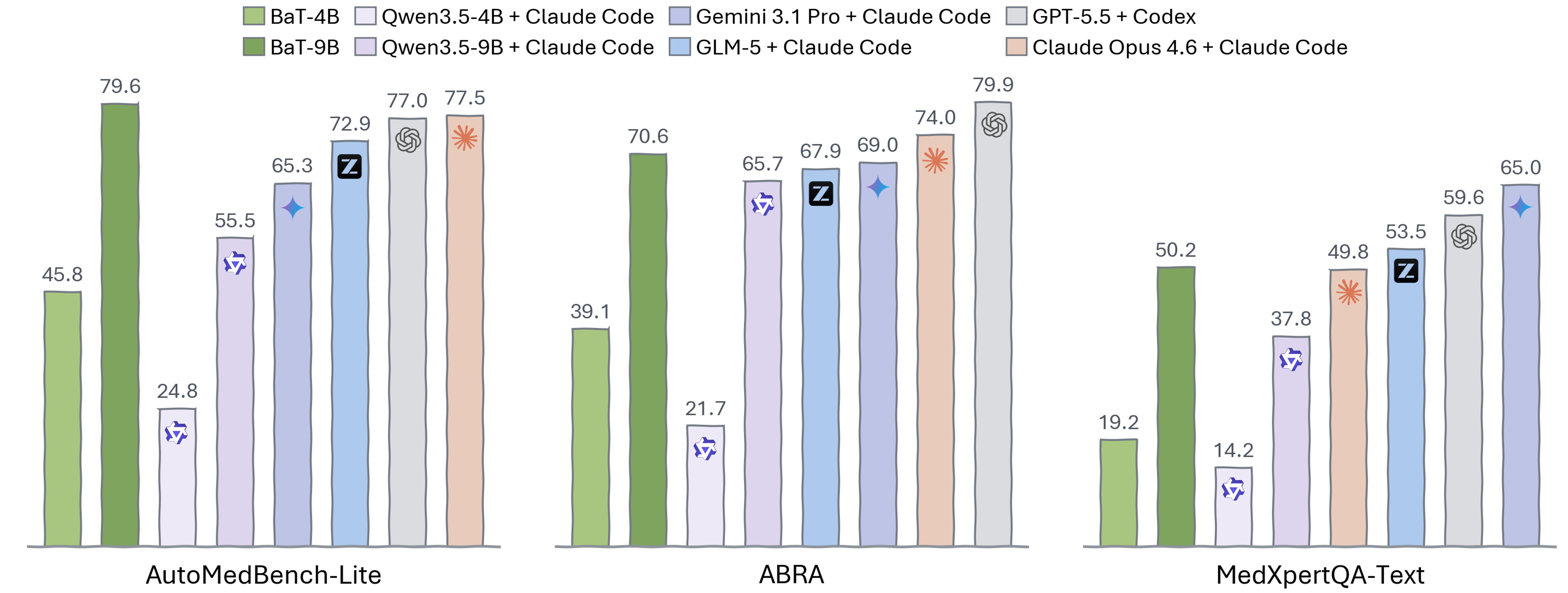}
    \caption{
        \textbf{\BaTNineBAgent{} leads AutoMedBench-Lite and trails the leaders by \BaTABRAGap{} points on ABRA and \BaTMedTextGap{} points on MedXpertQA-Text.}
        It scores \BaTAgentNineOverall{} on AutoMedBench-Lite, \BaTAgentNineABRA{} on ABRA, and \BaTAgentNineMedText{} on MedXpertQA-Text.
        The strongest non-BaT systems score \OpusAgentOverall{}, \ABRALeader{}, and \MedTextLeader{}, respectively.
        The legend shortens \BaTFourBAgent{} and \BaTNineBAgent{} to BaT-4B and BaT-9B.
        Each panel uses its benchmark's own protocol, and each system keeps the execution setting shown in the legend~\citep{liu2026automedbench,maksudov2026abra,zuo2025medxpertqa,qwen2026qwen354b,qwen2026qwen359b,google2026gemini31pro,glm2026glm5,openai2026gpt55,anthropic2026opus46}.
    }
    \label{fig:teaser}
\end{figure}

We evaluate medical performance on AutoMedBench-Lite, ABRA, and MedXpertQA-Text, and test transfer on eight external benchmarks~\citep{liu2026automedbench,maksudov2026abra,zuo2025medxpertqa}.
\BaTFourBModel{} and \BaTNineBModel{} more than double their Qwen Instruct Overall scores.
The full Stage Bank mixture leads every evaluated partial mixture.
\BaTNineBAgent{} reaches \BaTAgentNineOverall{} Overall, \BaTOpusMargin{} points above Claude Opus 4.6 with Claude Code (Figure~\ref{fig:teaser}).
On the external suite, the 9B policy remains within 3.4--5.8 points of its baseline on three reasoning tasks and improves $\tau^{2}$-Bench, SWE-bench Verified, and Terminal Bench 2.0.

BaT makes the benchmark part of the training system while preserving a content boundary around held-out tasks.
Our work makes four contributions:
\begin{itemize}
    \item \textbf{Benchmark-as-Teacher.}
    BaT turns a structured benchmark into an RSI system that joins diagnosis, content-isolated practice, policy updates, and re-evaluation.
    \item \textbf{Stage Bank.}
    The asynchronous Stage Bank pipeline synthesizes leakage-checked tasks and exposes targeted, mixed-stage, and E2E surrogate training states.
    \item \textbf{BiCuRL.}
    BiCuRL couples an outer stage curriculum and checkpoint fallback with inner rubric-verified GRPO updates.
    \item \textbf{BaT Agents.}
    \BaTFourBModel{} and \BaTNineBModel{} more than double their Qwen Instruct baselines, and \BaTNineBAgent{} reaches \BaTAgentNineOverall{} Overall on AutoMedBench-Lite.
\end{itemize}

\section{Benchmark-as-Teacher}
\label{sec:method}

\begin{figure}[h]
    \centering
    \includegraphics[width=0.8\columnwidth]{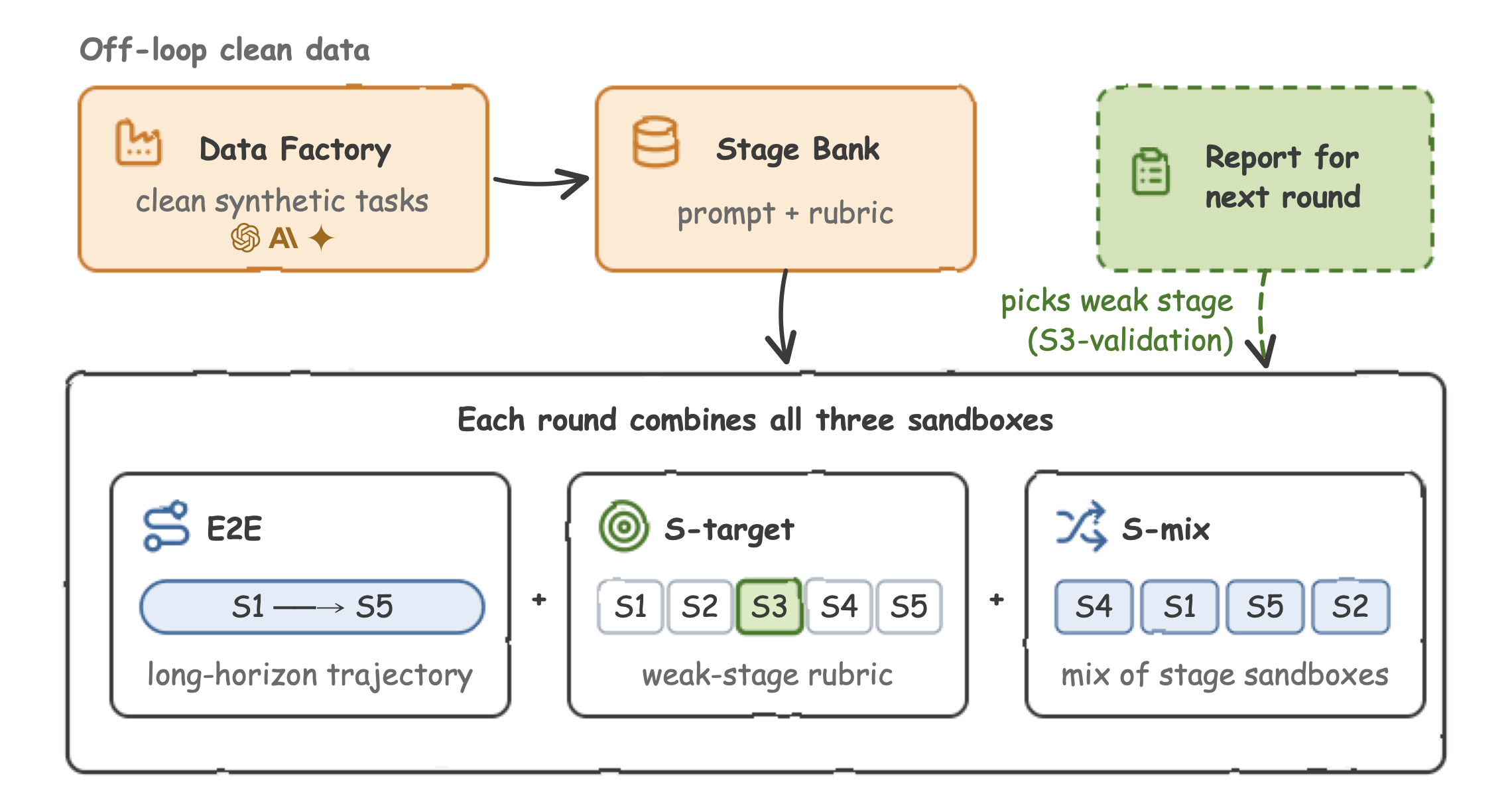}
    \caption{
        \textbf{Each BaT round combines all three sandboxes built from clean synthetic Stage Bank.}
        The Data Factory fills the Stage Bank, and the report for the next round only picks the weak stage:
        E2E trains on the whole unsegmented trajectory, S-target applies the weak-stage rubric, and S-mix blends stage sandboxes from the other stages.
        No evaluation task content enters training rows.
    }
    \label{fig:state-bank-views}
\end{figure}

Benchmark-as-Teacher is an RSI system with two coupled components.
The asynchronous Stage Bank pipeline prepares content-isolated practice states.
BiCuRL uses benchmark diagnostics to choose among those states, update the policy, retain checkpoints, and return the policy to evaluation.
Figure~\ref{fig:method-overview} shows this closed loop, and Figure~\ref{fig:state-bank-views} shows the Stage Bank pools.

\subsection{Stage Bank}
\label{sec:method-data}

\paragraph{Data Factory.}
Stage Bank builds practice data from public workflow descriptions and benchmark stage contracts.
We write a staged benchmark as
\begin{equation}
    \mathcal{B}=(\mathcal{S},\mathcal{C},\mathcal{V}),
\end{equation}
where $\mathcal{S}$ contains the ordered public stages and their boundaries, $\mathcal{C}$ contains rubric and evidence contracts, and $\mathcal{V}$ is a fixed held-out evaluation.
For policy $\pi$, the evaluation returns $\mathcal{V}(\pi)=(M,\mathbf{e})$, where $M$ is Overall and $\mathbf{e}$ contains one score per stage.
AutoMedBench-Lite defines five stages: Plan, Setup, Validate, Inference, and Submit~\citep{liu2026automedbench}.
Each stage and its rubric provide a shorter learning objective within the complete workflow.

The Data Factory synthesizes and validates all practice content outside the policy-update loop.
It asks teacher models to write fictional medical-imaging tasks and complete them as multi-turn trajectories.
Public workflow descriptions ground the task templates, while Self-Instruct and agent trajectory synthesis provide the data-generation pattern~\citep{liu2026healthagentbench,wang2023selfinstruct,xu2025agenttrek}.
A leakage preflight rejects held-out identifiers, paths, reports, traces, answers, and evaluation-derived metadata before a row enters training.
Aggregate recording statistics can guide synthesis, but raw evaluation content never enters a Stage Bank prompt or row.

\paragraph{SFT rows.}
Stage Bank turns each accepted teacher trajectory into single-response slices for the SFT cold start.
One row contains the task, stage skill, prior agent turns, and tool observations as context, followed by one teacher response as the training target.
This format keeps the multi-turn history while applying loss only to the selected response~\citep{xu2025agenttrek,chen2026agenticdpo}.
The resulting SFT data produce the initial policy $\theta_0$.

\paragraph{RL rows.}
Stage Bank builds RL data as executable sandboxes for full multi-turn rollouts~\citep{pan2024training,luo2025agentlightning}.
It reconstructs a fictional task at a public stage boundary and stores the resulting files, tools, and intermediate artifacts as the initial state.
A stage sandbox begins at one workflow boundary and carries that stage's goal, execution procedure, recovery steps, and rubric.
An End-to-End (E2E) surrogate chains all five stages in a smaller workflow that supports repeated rollouts.
We store SFT slices and RL sandbox rows separately; each RL row attaches an execution environment and a reward contract.
Appendix~\ref{sec:supp-stagebank} gives the data construction and E2E surrogate details, and Appendix~\ref{sec:supp-multiturn} gives the multi-turn RL objective.

Each verified Stage Bank row has the form
\begin{equation}
    z_i=(p_i,s_i,c_i,\mathcal{G}_i,\kappa_i,m_i),
\end{equation}
where $p_i$ is the sandbox state, $s_i$ names a stage or E2E, and $c_i$ contains rubric items.
$\mathcal{G}_i$ contains evidence requirements, $\kappa_i$ is the stage skill, and $m_i$ records provenance.
A stable \texttt{state\_id} links rollouts from the same row.
The bank indexes rows by task type, stage, outcome label, and reward source.

BiCuRL draws each round from three Stage Bank pools.
S-target contains states for the selected weak stage.
S-mix contains states from the remaining stages and limits forgetting.
E2E contains complete surrogate workflows and preserves cross-stage coordination.
The ablation changes only which pools enter training.

\begin{table}[t]   
    \caption{\textbf{SFT data (a) and RL Stage Bank data (b) across the five
    workflow stages and the E2E surrogate.}
    SFT rows select one response from a multi-turn teacher trajectory, while RL rows initialize executable sandboxes for multi-turn rollouts~\citep{xu2025agenttrek,chen2026agenticdpo,pan2024training,luo2025agentlightning}.
    We report average turns, average response tokens, and
    stored-row counts.}
    \centering
    {\scriptsize
    \setlength{\tabcolsep}{1.5pt}
    \renewcommand{\arraystretch}{1.08}
    \begin{minipage}[t]{0.47\columnwidth}
        \centering
        \textbf{(a) SFT}\\
        \resizebox{\linewidth}{!}{%
        \begin{tabular}{@{}lcccccc@{}}
            \toprule
            & \multicolumn{5}{c}{\textbf{S-target and S-mix}}
            & \multicolumn{1}{c}{\textbf{End-to-End}} \\
            \cmidrule(lr){2-6}\cmidrule(lr){7-7}
            \textbf{Statistic}
            & \multicolumn{1}{c}{\shortstack{\textbf{S1}\\Plan}}
            & \multicolumn{1}{c}{\shortstack{\textbf{S2}\\Setup}}
            & \multicolumn{1}{c}{\shortstack{\textbf{S3}\\Validate}}
            & \multicolumn{1}{c}{\shortstack{\textbf{S4}\\Infer}}
            & \multicolumn{1}{c}{\shortstack{\textbf{S5}\\Submit}}
            & \multicolumn{1}{c}{\shortstack{\textbf{E2E}\\surrogate}} \\
            \midrule
            Turns & 1.0 & 1.0 & 1.0 & 1.0 & 1.0 & 1.0 \\
            Tokens (K) & 0.4 & 1.4 & 0.3 & 0.3 & 0.3 & 0.4 \\
            Count (K) & 2.0 & 2.0 & 2.0 & 2.0 & 2.0 & 2.0 \\
            \bottomrule
        \end{tabular}}%
    \end{minipage}\hfill
    \begin{minipage}[t]{0.47\columnwidth}
        \centering
        \textbf{(b) RL}\\
        \resizebox{\linewidth}{!}{%
        \begin{tabular}{@{}lcccccc@{}}
            \toprule
            & \multicolumn{5}{c}{\textbf{S-target and S-mix}}
            & \multicolumn{1}{c}{\textbf{End-to-End}} \\
            \cmidrule(lr){2-6}\cmidrule(lr){7-7}
            \textbf{Statistic}
            & \multicolumn{1}{c}{\shortstack{\textbf{S1}\\Plan}}
            & \multicolumn{1}{c}{\shortstack{\textbf{S2}\\Setup}}
            & \multicolumn{1}{c}{\shortstack{\textbf{S3}\\Validate}}
            & \multicolumn{1}{c}{\shortstack{\textbf{S4}\\Infer}}
            & \multicolumn{1}{c}{\shortstack{\textbf{S5}\\Submit}}
            & \multicolumn{1}{c}{\shortstack{\textbf{E2E}\\surrogate}} \\
            \midrule
            Turns & 25.6 & 40.5 & 36.7 & 37.8 & 28.5 & 45.5 \\
            Tokens (K) & 8.4 & 11.0 & 10.4 & 10.8 & 8.5 & 12.1 \\
            Count (K) & 3.7 & 6.9 & 5.5 & 2.9 & 1.4 & 1.2 \\
            \bottomrule
        \end{tabular}}%
    \end{minipage}
    }
    \label{tab:training-data}
\end{table}

\subsection{Bilevel Curriculum Reinforcement Learning (BiCuRL)}
\label{sec:method-loop}

BiCuRL is the self-improving post-training method inside BaT.
Its outer loop chooses what to practice from held-out stage scores, and its inner loop updates the policy on that curriculum with GRPO~\citep{shao2024deepseekmath}.
At round $r$, the fixed controller evaluation returns an Overall score $M_r$ and stage scores $\mathbf{e}_r=(e_{r,s})_{s\in\mathcal{S}}$.
Over the Stage Bank mixture family $\mathcal{Q}_{\mathrm{SB}}$, BiCuRL targets the bilevel objective
\begin{equation}
    \begin{aligned}
        \max_{\mathbf{q}\in\mathcal{Q}_{\mathrm{SB}}}\quad
        &\mathcal{M}\left(\hat{\theta}(\mathbf{q})\right),\\
        \text{s.t.}\quad
        &\hat{\theta}(\mathbf{q})\in
        \arg\max_{\theta}
        \mathcal{J}_{\mathrm{GRPO}}(\theta;\mathbf{q}).
    \end{aligned}
    \label{eq:bicurl-bilevel}
\end{equation}
BiCuRL approximates this objective through alternating updates.
Each round runs a finite GRPO block, evaluates the candidate, updates the curriculum, and starts the next block from the retained checkpoint.
The controller evaluation and final evaluation use disjoint runs.
Only five stage scores and one Overall score leave the controller evaluation; the controller scores and discards its rollouts.

\subsubsection{Inner Loop: Rubric-Verified Agentic RL}

Given curriculum $\mathbf{q}_r$, BiCuRL samples Stage Bank rows $z_i$ and draws $K$ rollouts from policy $\pi_{\bar{\theta}_r}$ in sandbox state $p_i$.
An LLM rubric verifier scores rollout $y_{i,k}$ against rubric items $c_i$ using execution records and artifact evidence $x_{i,k}$~\citep{zheng2023llmjudge}.
The verifier also reports evidence completeness $\eta_{i,k}\in[0,1]$, the fraction of required evidence confirmed by the rollout and its artifacts:
\begin{equation}
    \begin{aligned}
        v_{i,k,\ell}
        &=\operatorname{Verify}
        (\ell;p_i,y_{i,k},x_{i,k})\in\{0,1\},\\
        r_{i,k}
        &=\eta_{i,k}
        \frac{1}{|c_i|}\sum_{\ell\in c_i}v_{i,k,\ell},\\
        A_{i,k}
        &=\frac{r_{i,k}-K^{-1}\sum_{j=1}^{K}r_{i,j}}
        {\operatorname{std}_{j}(r_{i,j})+\epsilon_{\mathrm{adv}}}.
    \end{aligned}
    \label{eq:bicurl-reward}
\end{equation}
Binary rubric decisions avoid free-form score calibration, while evidence completeness lowers rewards for unsupported success claims.
Group normalization compares rollouts that share one state and reward contract.
GRPO uses the resulting advantages to produce candidate checkpoint $\widetilde{\theta}_{r+1}$.

\subsubsection{Outer Loop: Stage Routing and Checkpoint Fallback}

The curriculum router reads $\mathbf{e}_r$ and writes an auditable report that selects target stage $s_r^\star$.
The round curriculum mixes three Stage Bank pools with fixed proportions $\boldsymbol{\rho}$:
\begin{equation}
    \begin{aligned}
        q_r(z)
        ={}&\rho_{\mathrm{target}}\,q_{\mathrm{target}}(z\mid s_r^\star)
        +\rho_{\mathrm{mix}}\,q_{\mathrm{mix}}(z\mid s_r^\star)\\
        &+\rho_{\mathrm{E2E}}\,q_{\mathrm{E2E}}(z).
    \end{aligned}
    \label{eq:bicurl-curriculum}
\end{equation}
S-target changes with the selected stage, while S-mix and E2E preserve the rest of the workflow.

The router also controls fallback to the best retained checkpoint $\theta^\star$.
A counter $c_r$ records consecutive score drops, and fallback fires after three drops or a policy shift above threshold $\tau$:
\begin{equation}
    \begin{aligned}
        c_{r+1}&=
        \begin{cases}
            c_r+1, & M_{r+1}<M_r,\\
            0, & \text{otherwise},
        \end{cases}\\
        \bar{\theta}_{r+1}&=
        \begin{cases}
            \theta^\star, & c_{r+1}\geq 3
            \;\text{or}\;
            D_{\mathrm{KL}}\!\left(\pi_{\widetilde{\theta}_{r+1}}\,\|\,\pi_{\theta^\star}\right)>\tau,\\
            \widetilde{\theta}_{r+1}, & \text{otherwise}.
        \end{cases}
    \end{aligned}
    \label{eq:bicurl-update}
\end{equation}
BaT records each round's Stage Bank states, pool mixture, reward version, controller scores, and retained checkpoint.
The next round starts only after every selected state passes the leakage preflight.

\subsection{Agent}
A BaT Agent combines a BiCuRL-trained policy with a fixed \BaTAgentRuntime{} execution environment~\citep{wang2025openhands}.
This environment includes public stage skills distilled from the training-time LLM prescriptions.
The skills restate each stage's goal, checks, and recovery steps without carrying evaluation content.
Inside \BaTAgentRuntime{}, the agent can edit files, run commands, inspect results, and submit artifacts.
We keep this engineering layer fixed across BaT Agent comparisons and exclude it from BiCuRL: the policy is the only part that training changes.

\section{Experimental Setting}
\label{sec:setup}

\subsection{Benchmark and Metrics}
\paragraph{Medical benchmarks.}
AutoMedBench is a long-horizon benchmark for medical-AI research agents~\citep{liu2026automedbench}.
It scores five workflow stages across two difficulty tiers.
We use the AutoMedBench-Lite tier, which provides more task-brief support while keeping the same workflow and scoring structure.
Our diagnostic evaluation suite contains \BaTEvaluationTracks{} long-horizon task tracks.
For each track, we repeat the task \BaTEvaluationRepeats{} times, giving \BaTEvaluationCells{} runs per evaluated system.
The tested agents average \BaTAverageTurns{} interaction turns per run.

ABRA tests radiology agents inside an OHIF viewer and an Orthanc DICOM server~\citep{maksudov2026abra}.
Its 655 tasks span three difficulty tiers and eight task types, and agents use 21 tools for image navigation, annotation, and reporting.
MedXpertQA tests expert medical knowledge and reasoning across 17 specialties and 11 body systems~\citep{zuo2025medxpertqa}.
We use its Text subset, which contains text-only medical questions.
AutoMedBench-Lite and ABRA evaluate full agent systems, while MedXpertQA-Text evaluates text responses.
Figure~\ref{fig:teaser} keeps the system setting shown in its legend and supports comparisons within each benchmark panel.
Its values are the recorded evaluation aggregates for those system settings under each benchmark's native scorer.
The cited benchmark papers define the tasks and scorers; our evaluations supply the plotted model scores.

\paragraph{AutoMedBench-Lite metrics.}
AutoMedBench-Lite measures both process and outcome.
\textbf{Task} scores the submitted result.
\textbf{Agentic} scores completion of Plan, Setup, Validate, Inference, and Submit.
\textbf{Overall} gives Task and Agentic equal weight and uses their unrounded values.
We use Overall as the main summary because a long-horizon agent must complete the workflow and produce a valid result.
We report Task and Agentic separately to show whether a change comes from the process or the outcome.
We report all three scores on a 0--100 scale with one decimal.
We compute score differences from unrounded values and then round each difference to one decimal.
The statistical unit is the task track.
We average the ten repeats within each track and then average the seven track means.
This two-level aggregation avoids treating all 70 runs as independent.
Table~\ref{tab:automedbench-results} carries over the source reported uncertainty half-widths across the seven track-level means.

\subsection{Models and Baselines}

We train the Instruct versions of Qwen3.5-4B and Qwen3.5-9B~\citep{qwen2026qwen354b,qwen2026qwen359b}.
At each size, Figure~\ref{fig:bat-vs-grpo} compares the Qwen Instruct Baseline, supervised fine-tuning (SFT), GRPO, and BiCuRL inside the full BaT loop~\citep{ouyang2022training,shao2024deepseekmath}.
Baseline always denotes the corresponding Qwen Instruct checkpoint before post-training.
SFT reports the cold-start checkpoint alone, GRPO applies group-relative policy optimization with a single final task reward on E2E data, and BiCuRL applies stage-guided post-training after the same SFT cold start.

Figure~\ref{fig:teaser} compares systems on AutoMedBench-Lite, ABRA, and MedXpertQA-Text~\citep{liu2026automedbench,maksudov2026abra,zuo2025medxpertqa}.
The comparison includes Qwen3.5-4B and Qwen3.5-9B with Claude Code, Gemini 3.1 Pro with Claude Code, and GLM-5 with Claude Code~\citep{qwen2026qwen354b,qwen2026qwen359b,google2026gemini31pro,glm2026glm5}.
It also includes GPT-5.5 with Codex and Claude Opus 4.6 with Claude Code~\citep{openai2026gpt55,anthropic2026opus46}.
Each system keeps the execution setting named in the figure.
Figure~\ref{fig:bat-vs-grpo} reports policy-level post-training runs, while Figure~\ref{fig:teaser} reports full agent-system runs.

\subsection{Training Data and Checks}

Stage Bank contains \BaTStateBankRows{} candidate prompt states that support rubric scoring.
A separate content-isolated synthetic E2E source pool contains \BaTCleanEERows{} rows.
Stage Bank projects that source pool into E2E, S-target, and S-mix sandboxes without copying evaluation content.
The 4B and 9B SFT starts use \BaTFourBSFTRows{} and \BaTNineBSFTRows{} rows.
Each GRPO group samples \BaTRolloutsPerPrompt{} continuations from one state.
We prepared a \BaTMatchedAblationRows{}-row matched ablation pool.
Generation-time rules normalize required fields and block known task and path markers.
A hard preflight covers every training and validation file, and the rubric judge scores fresh continuations against each row's rubric.
We keep the diagnostic evaluation fixed across rounds and apply the leakage rules described above before every update.
We will release the source manifests, synthesis prompts, checker settings, pool mixtures, optimizer settings, and leakage rules with the training data.

\section{Results}
\label{sec:results}

The results connect the BaT system to three claims.
The Overall scores of \BaTFourBModel{} and \BaTNineBModel{} more than double their Qwen Instruct baselines, \BaTNineBAgent{} leads AutoMedBench-Lite and approaches the leaders on ABRA and MedXpertQA-Text, and external transfer depends on model size.
Figure~\ref{fig:bat-vs-grpo} scores trained policies, while Figure~\ref{fig:teaser} compares medical benchmark results.
We report these levels separately and avoid direct comparisons between them.

\subsection{BiCuRL More Than Doubles Baselines}
\label{sec:results-scale}

\begin{figure}[t]
    \centering
    \includegraphics[width=\columnwidth]{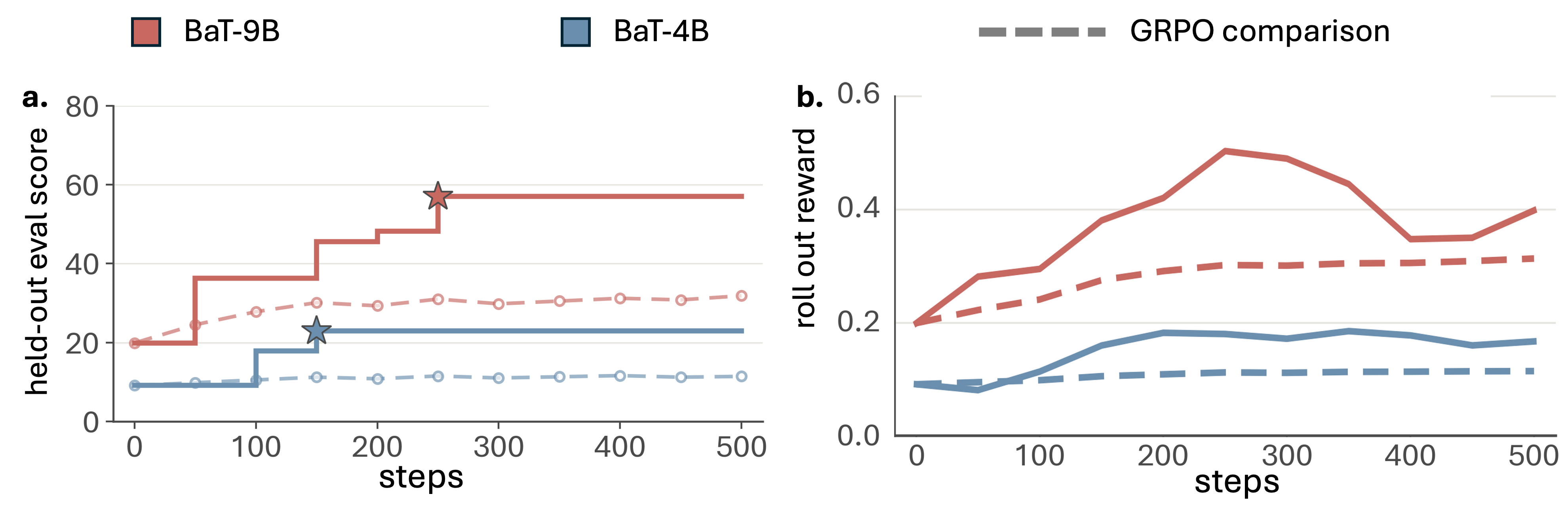}
    \caption{
        \textbf{BaT keeps improving across rounds, while GRPO with one final reward saturates early.}
        Solid step lines track the best BaT checkpoint so far and stars mark the selected checkpoints; dashed curves show GRPO in the matching model color.
    }
    \label{fig:bat-vs-grpo}
\end{figure}

Table~\ref{tab:automedbench-results} and Figure~\ref{fig:bat-vs-grpo} report the completed \BaTFourBModel{} and \BaTNineBModel{} policy runs.
\BaTFourBModel{} has 22.9 Overall, compared with 6.1 for the Instruct baseline.
\BaTNineBModel{} has 53.4 Overall, compared with 19.9 for the Instruct baseline.
The 9B score is also 21.5 points above GRPO after rounding the displayed means.
The figure keeps the recorded protocol for each completed run, so these differences describe the observed scores instead of a matched training ablation.

\begin{table}[t]
    \caption{
        \textbf{BiCuRL more than doubles the Qwen Instruct AutoMedBench-Lite Overall score at both model sizes.}
        Each score is the mean over the seven track-level means, with $\pm$ showing the source draft's reported uncertainty half-width.
        Bold and underline mark the best and second-best result within each model size.
        Baseline denotes Qwen3.5 Instruct~\citep{qwen2026qwen354b,qwen2026qwen359b}.
        SFT follows supervised instruction tuning~\citep{ouyang2022training}, and GRPO follows group-relative policy optimization~\citep{shao2024deepseekmath}.
        Ovl. denotes Overall, and Agt. denotes Agentic.
    }
    \centering
    \newcommand{\cipm}[2]{#1\,{\scriptsize\textit{$\pm$#2}}}
    {\scriptsize
    \setlength{\tabcolsep}{5pt}
    \renewcommand{\arraystretch}{1.12}
    \begin{tabular}{@{}lcccccc@{}}
        \toprule
        & \multicolumn{3}{c}{\textbf{Qwen3.5-4B}~\citep{qwen2026qwen354b}}
        & \multicolumn{3}{c}{\textbf{Qwen3.5-9B}~\citep{qwen2026qwen359b}} \\
        \cmidrule(lr){2-4}\cmidrule(lr){5-7}
        \textbf{Training}
        & \textbf{Ovl.} & \textbf{Agt.} & \textbf{Task}
        & \textbf{Ovl.} & \textbf{Agt.} & \textbf{Task} \\
        \midrule
        Baseline
        & \cipm{6.1}{1.7} & \cipm{12.1}{2.0} & \cipm{0.0}{1.1}
        & \cipm{19.9}{2.3} & \cipm{11.3}{2.0} & \cipm{\underline{28.4}}{2.6} \\
        SFT~\citeyearpar{ouyang2022training}
        & \cipm{\underline{18.1}}{2.3} & \cipm{10.8}{1.9} & \cipm{\textbf{25.4}}{2.5}
        & \cipm{12.9}{2.1} & \cipm{21.6}{2.4} & \cipm{4.1}{1.5} \\
        GRPO~\citeyearpar{shao2024deepseekmath}
        & \cipm{11.4}{2.0} & \cipm{\underline{15.9}}{2.2} & \cipm{6.9}{1.7}
        & \cipm{\underline{31.9}}{2.7} & \cipm{\underline{39.2}}{2.8} & \cipm{24.5}{2.5} \\
        \rowcolor{batrowgreen}
        BiCuRL
        & \cipm{\textbf{22.9}}{2.4} & \cipm{\textbf{28.8}}{2.6} & \cipm{\underline{17.0}}{2.2}
        & \cipm{\textbf{53.4}}{3.2} & \cipm{\textbf{64.7}}{3.1} & \cipm{\textbf{42.1}}{3.2} \\
        \bottomrule
    \end{tabular}
    }
    \label{tab:automedbench-results}
\end{table}

\textbf{Takeaway.}
\BaTFourBModel{} and \BaTNineBModel{} reach 22.9 and 53.4 Overall, more than twice their corresponding Instruct baselines.

\subsection{BiCuRL Performance across Training Rounds}
\label{sec:results-evolution}

BiCuRL repeats evaluation, stage selection, and post-training over several rounds.
Figure~\ref{fig:method-overview}(c) reports how the Overall score changes across rounds for both model sizes.
At both sizes, the retained BiCuRL checkpoint passes the corresponding GRPO score early in training.
The best-so-far staircases preserve each accepted gain through round ten even when a later candidate scores lower.
The raw round curves fluctuate, which motivates checkpoint retention and fallback in the outer loop.

\textbf{Takeaway.}
Some candidates score lower than their predecessors; checkpoint retention preserves the best observed Overall score across later rounds.

\subsection{BiCuRL Ablation Study}
\label{sec:results-ablation}

\begin{figure}[!htbp]
    \centering
    \includegraphics[width=0.6\columnwidth]{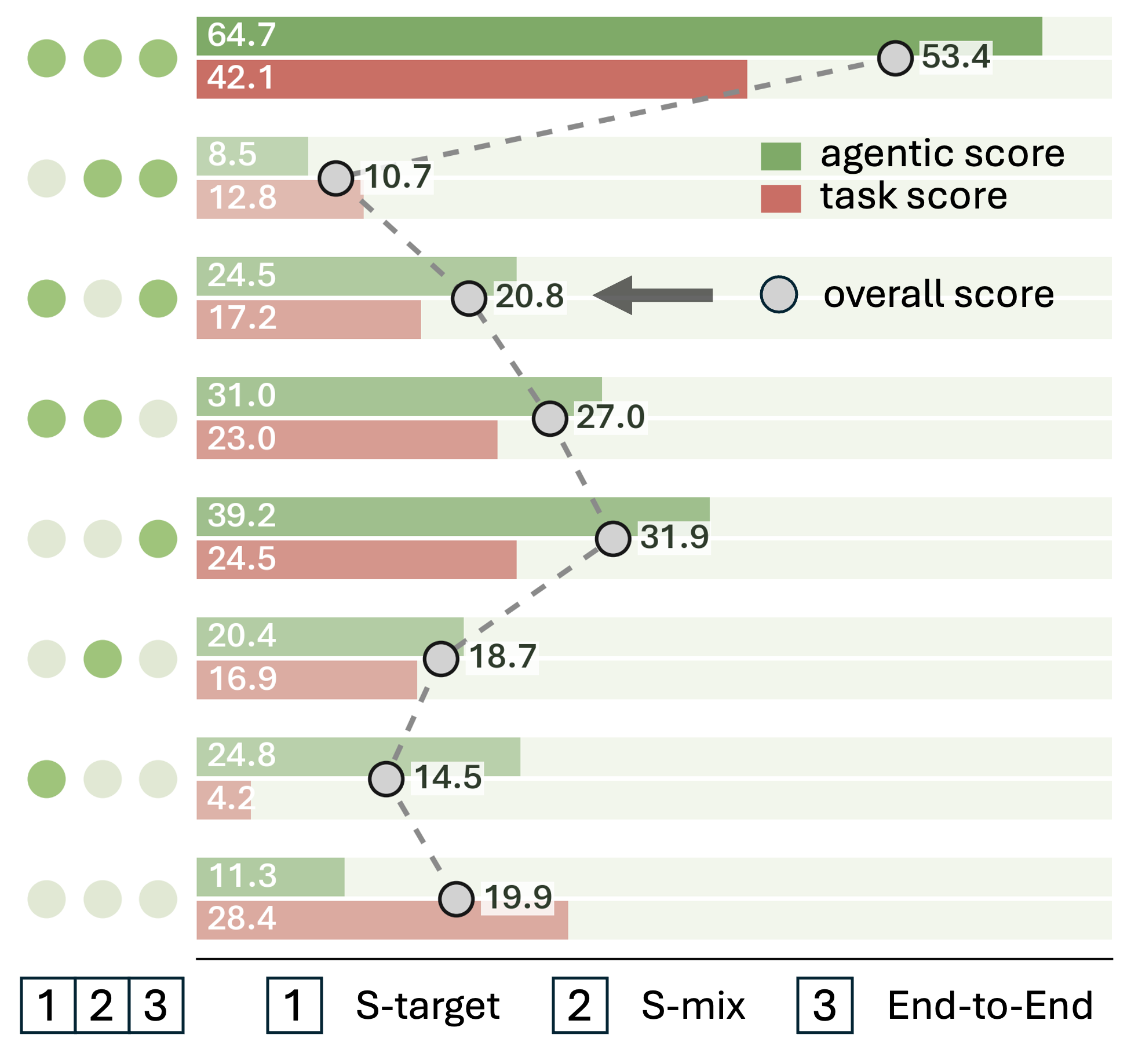}
    \caption{
        \textbf{The full S-target, S-mix, and E2E sandbox mix leads every partial mix.}
        The dot matrix on the left marks which pools enter training (columns 1--3: S-target, S-mix, End-to-End; dark included, pale omitted).
        Green bars show Agentic, red bars show Task, and gray circles show Overall.
        Rows from top to bottom: full BaT, E2E+S-mix, E2E+S-target, S-target+S-mix, E2E, S-mix, S-target, and the Qwen3.5-9B Instruct baseline.
        All scores use a 0--100 scale and one decimal.
    }
    \label{fig:ablation-study}
\end{figure}

Figure~\ref{fig:ablation-study} compares the S-target, S-mix, and E2E pools defined in Section~\ref{sec:method-data}.
The full three-pool run leads with 53.4 Overall, matching the \BaTNineBModel{} row in Figure~\ref{fig:bat-vs-grpo}.
The strongest partial mix, E2E alone, reaches 31.9, a gap of 21.5 points.
Among the three runs that each omit one pool, every point estimate trails the full mix by at least 26 points.

\textbf{Takeaway.}
The full sandbox mix leads every partial mix by at least 21.5 Overall points and each drop-one-pool run by at least 26 points.

\subsection{BaT-9B Ranks First among Local LLMs}
\label{sec:results-local-llms}

\begin{figure*}[!htbp]
    \centering
    \includegraphics[width=\linewidth]{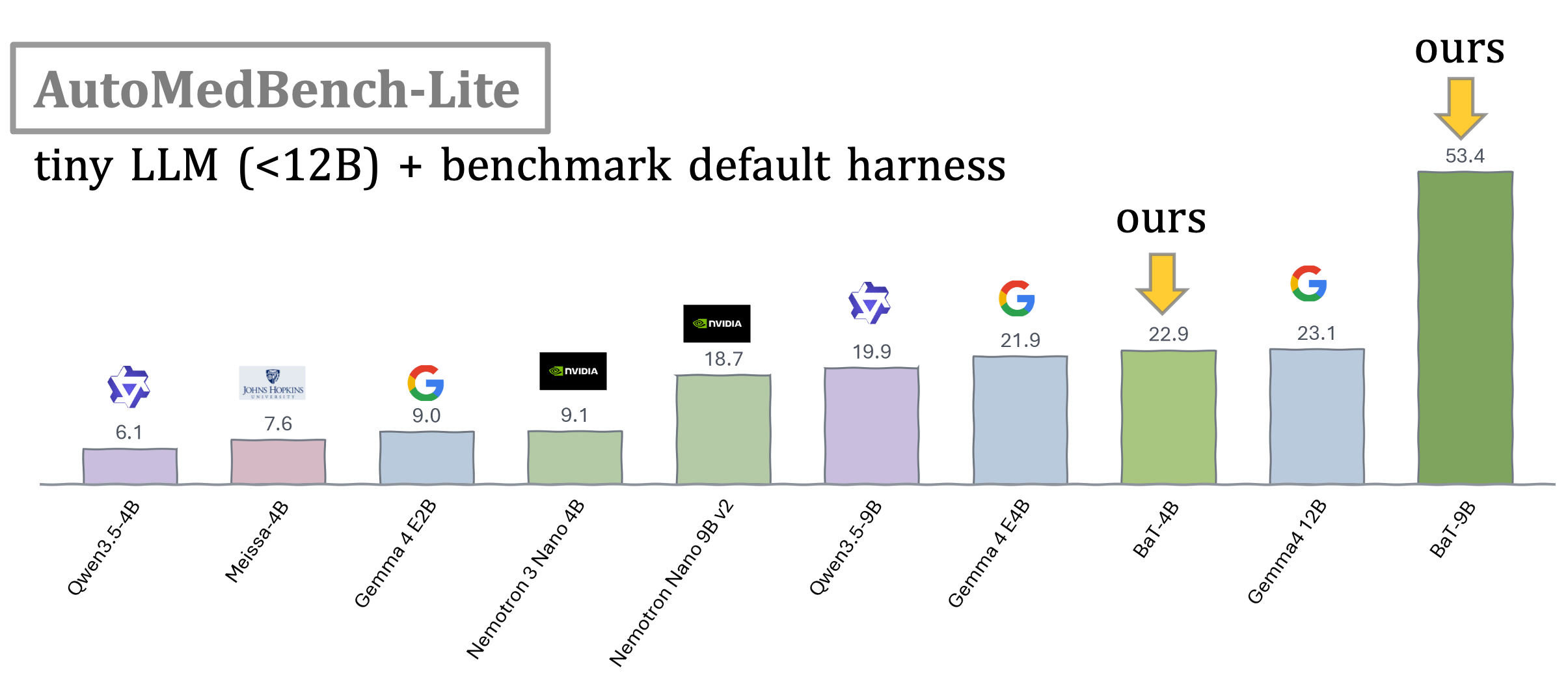}
    \caption{
        \textbf{\BaTNineBModel{} ranks first, and \BaTFourBModel{} ranks third among 10 tiny local LLMs under the same default runner.}
        Here tiny local LLM is defined as parameter size no more than 12B.
        Bars report AutoMedBench-Lite Overall scores.
    }
    \label{fig:local-llm-comparison}
\end{figure*}

Given the critical privacy requirements in medical research workflows, local deployment is highly desirable. Tiny models, defined as those with fewer than 12B parameters, offer a cost-effective and performant alternative, prompting us to download and test a range of representative open-weight models.
\figureautorefname~\ref{fig:local-llm-comparison} compares tiny local LLMs with the same default \BaTAgentRuntime{} runner and benchmark protocol~\citep{wang2025openhands}.
\BaTFourBModel{} raises its Qwen3.5-4B backbone from 6.1 to 22.9 Overall.
\BaTNineBModel{} ranks first at 53.4 Overall, leading 2x compared to the second place Gemma 12B~\citep{gemmateam2026gemma4}.
For further comparison with top-tier open-source local LLMs, see \tableautorefname~\ref{tab:local-llm-results}.

\textbf{Takeaway.}
BaT moves both Qwen backbones upward, and the 4B policy trails only one larger-total-parameter LLM.

\subsection{BaT-9B Leads One Medical Benchmark and Approaches Two Leaders}
\label{sec:results-agent}

Figure~\ref{fig:teaser} compares systems on three medical benchmarks~\citep{liu2026automedbench,maksudov2026abra,zuo2025medxpertqa}.
Each BaT Agent pairs its BiCuRL-trained policy with the fixed \BaTAgentRuntime{} environment described under Agent.
On AutoMedBench-Lite, \BaTNineBAgent{} reaches \BaTAgentNineOverall{}, \BaTOpusMargin{} points above Claude Opus 4.6 with Claude Code at \OpusAgentOverall{}~\citep{anthropic2026opus46}.
For AutoMedBench-Lite per-track details, \figureautorefname~\ref{fig:downstream-score-tracks} in Appendix~\ref{sec:supp-protocol} presents a preliminary per-track comparison of \BaTFourBAgent{}, \BaTNineBAgent{}, and Claude Opus 4.6 with Claude Code.
On ABRA, it reaches \BaTAgentNineABRA{}, compared with \ABRALeader{} for GPT-5.5 with Codex, a \BaTABRAGap{}-point gap~\citep{openai2026gpt55}.
On MedXpertQA-Text, it reaches \BaTAgentNineMedText{}, compared with \MedTextLeader{} for Gemini 3.1 Pro with Claude Code, a \BaTMedTextGap{}-point gap~\citep{google2026gemini31pro}.
The 4B BaT Agent scores \BaTAgentFourAutoMed{}, \BaTAgentFourABRA{}, and \BaTAgentFourMedText{} on the same three benchmarks.
At both sizes, each BaT Agent outperforms Qwen3.5 with Claude Code on all three benchmarks~\citep{qwen2026qwen354b,qwen2026qwen359b}.

\textbf{Takeaway.}
\BaTNineBAgent{} leads AutoMedBench-Lite and trails the leaders by \BaTABRAGap{} points on ABRA and \BaTMedTextGap{} points on MedXpertQA-Text.

\subsection{External Benchmarks Show Scale-Dependent Retention}
\label{sec:results-external}

Strong medical results could come with weaker general reasoning or tool use.
Table~\ref{tab:external-results} compares each BaT policy with its Instruct baseline on three short-turn reasoning tasks and five long-horizon tasks~\citep{maa2026aime,rein2023gpqa,barres2025tau2bench,patil2025bfcl,mialon2024gaia,openai2024swebenchverified,merrill2026terminalbench}.
The 9B policy remains within 3.4--5.8 points of its baseline on the three reasoning benchmarks and improves $\tau^{2}$-Bench by 5.4 points.
It also improves SWE-bench Verified and Terminal Bench 2.0, while BFCL-Parity and GAIA remain below the baseline.
The 4B policy scores below its baseline on all eight benchmarks.
The pattern links transfer to model size and task type.

\begin{table}[t]
    \caption{
        \textbf{BaT-9B retains most reasoning accuracy and improves three of five long-horizon scores, while BaT-4B declines on all eight benchmarks.}
        Base denotes each Qwen3.5 Instruct checkpoint~\citep{qwen2026qwen354b,qwen2026qwen359b}.
        Delta is BaT minus Base in percentage points; green marks gains and red marks drops.
        Each pair uses the same benchmark protocol.
    }
    \centering
    {\scriptsize
    \setlength{\tabcolsep}{3pt}
    \renewcommand{\arraystretch}{1.10}
    \begin{tabular}{@{}lrrrrrr@{}}
        \toprule
        & \multicolumn{3}{c}{\textit{Qwen3.5-4B}~\citep{qwen2026qwen354b}}
        & \multicolumn{3}{c}{\textit{Qwen3.5-9B}~\citep{qwen2026qwen359b}} \\
        \cmidrule(lr){2-4}\cmidrule(lr){5-7}
        \textbf{Benchmark}
        & \textbf{Base} & \textbf{BaT} & $\boldsymbol{\Delta}$
        & \textbf{Base} & \textbf{BaT} & $\boldsymbol{\Delta}$ \\
        \midrule
        \multicolumn{7}{@{}l}{\textit{short-turn reasoning tasks}} \\
        AIME 2025~\citep{maa2026aime}
        & 85.0 & 56.7 & \textcolor{deltaNegative}{\textbf{-28.3}}
        & 88.3 & 83.6 & \textcolor{deltaNegative}{\textbf{-4.7}} \\
        AIME 2026~\citep{maa2026aime}
        & 89.7 & 53.3 & \textcolor{deltaNegative}{\textbf{-36.4}}
        & 92.5 & 86.7 & \textcolor{deltaNegative}{\textbf{-5.8}} \\
        GPQA-Diamond~\citep{rein2023gpqa}
        & 76.2 & 58.1 & \textcolor{deltaNegative}{\textbf{-18.1}}
        & 81.7 & 78.3 & \textcolor{deltaNegative}{\textbf{-3.4}} \\
        \midrule
        \multicolumn{7}{@{}l}{\textit{long-horizon tasks}} \\
        $\tau^{2}$-Bench~\citep{barres2025tau2bench}
        & 79.9 & 29.1 & \textcolor{deltaNegative}{\textbf{-50.8}}
        & 79.1 & 84.5 & \textcolor{deltaPositive}{\textbf{+5.4}} \\
        BFCL-Parity~\citep{patil2025bfcl}
        & 62.6 & 10.6 & \textcolor{deltaNegative}{\textbf{-52.0}}
        & 79.4 & 68.3 & \textcolor{deltaNegative}{\textbf{-11.1}} \\
        GAIA~\citep{mialon2024gaia}
        & 15.2 & 2.4 & \textcolor{deltaNegative}{\textbf{-12.8}}
        & 24.9 & 21.2 & \textcolor{deltaNegative}{\textbf{-3.7}} \\
        SWE-bench Verified~\citep{openai2024swebenchverified}
        & 39.4 & 4.8 & \textcolor{deltaNegative}{\textbf{-34.6}}
        & 49.0 & 51.8 & \textcolor{deltaPositive}{\textbf{+2.8}} \\
        Terminal Bench 2.0~\citep{merrill2026terminalbench}
        & 11.2 & 1.1 & \textcolor{deltaNegative}{\textbf{-10.1}}
        & 19.9 & 21.3 & \textcolor{deltaPositive}{\textbf{+1.4}} \\
        \bottomrule
    \end{tabular}
    }
    \label{tab:external-results}
\end{table}

\textbf{Takeaway.}
BaT-9B stays close to its short-turn reasoning baseline and improves three long-horizon scores; BaT-4B scores lower on all eight tests.

\section{Discussion}
\label{sec:discussion}

\subsection{Why the Loop Matters}

BaT uses a benchmark in two linked roles.
AutoMedBench-Lite measures the current agent, and its aggregate diagnostics control what the model practices next.
Its task content stays outside training rows.
Its scores set later training mixtures, which makes the suite a diagnostic evaluation and rules out treating it as an untouched final test.
External benchmarks provide a separate transfer check, although their protocols differ.
BaT requires benchmarks that expose stable stages, stage-level scores, and rubrics or checks that can score a continuation.
The separation between Stage Bank and BiCuRL matters operationally: synthesis and validation can run asynchronously, while the policy-update loop reads only versioned, executable states.

\subsection{Limits}

The current evidence has six limits.
First, AutoMedBench-Lite supplies aggregate diagnostic signals, so the suite participates in adaptation even though task content stays outside training rows.
This setup provides content isolation, while a separate untouched final test and a measured semantic leakage audit remain future work.
Second, the external table combines retained local runs with published baselines, and its local aggregates lack complete per-run manifests.
Third, \BaTFourBModel{} scores below its baseline on all eight external benchmarks, while \BaTNineBModel{} gains on three of five long-horizon tasks.
Fourth, the top-tier comparison evaluates complete agent systems with their named execution settings, so it supports only a system-level claim.

\section{Related Work}
\label{sec:related-work}

\subsection{Medical and Long-Horizon Agent Benchmarks}

Long-horizon benchmarks test agents through tools, files, and many-step tasks.
Terminal Bench 2.0 tests command-line work in isolated environments~\citep{merrill2026terminalbench}.
The Berkeley Function Calling Leaderboard (BFCL) focuses on function calling~\citep{patil2025bfcl}.
IFBench tests unseen instruction rules~\citep{pyatkin2025ifbench}.
In healthcare, MedAgentBench evaluates physician-authored EHR tasks in a FHIR environment, HealthAgentBench spans realistic agentic healthcare settings, and AutoMedBench adds stage scores for medical AI development~\citep{jiang2025medagentbench,liu2026healthagentbench,liu2026automedbench}.
These benchmarks measure agent performance, while the stage structure in AutoMedBench also supports diagnosis.
BaT uses public stage structure to build content-isolated training sandboxes and choose the focus of the next round.

\subsection{Agent Post-Training}

GRPO compares samples from the same group and avoids a learned value model~\citep{shao2024deepseekmath}.
ReAct joins reasoning with tool actions~\citep{yao2023react}.
Process supervision assigns feedback to intermediate reasoning steps, while curriculum learning schedules training examples by difficulty or structure~\citep{lightman2023verify,bengio2009curriculum}.
Self-Refine uses model feedback to revise outputs during inference~\citep{madaan2023selfrefine}.
Agent Lightning converts multi-step agent traces into RL transitions with credit assignment~\citep{luo2025agentlightning}.
Self-Rewarding Language Models and SPIN use iterative model-generated feedback or self-play to improve later checkpoints~\citep{yuan2024selfrewarding,chen2024spin}.
Frontis-MA1 develops recursive self-improvement for machine-learning engineering through OpenMLE-Gym, OpenMLE-RL, and OpenMLE-Evo~\citep{yang2026frontisma1}.
BaT instead centers a structured held-out benchmark in the control loop.
Public stages define practice targets, public rubrics score rollouts, and aggregate diagnostics change the next round's Stage Bank mixture.

\subsection{Execution Environments and Synthetic Task Generation}

OpenHands provides a workspace for agents that edit files, run programs, and inspect results~\citep{wang2025openhands}.
SWE-World studies software-agent training without full Docker execution~\citep{sun2026sweworld}.
SWE-Gym supplies executable software-engineering environments for training agents and verifiers, while VerlTool provides modular tool-use RL infrastructure with asynchronous rollout support~\citep{pan2024training,jiang2025verltool}.
On the data side, Self-Instruct bootstraps instruction data from model-written tasks, and later agent pipelines scale this recipe to tool-use trajectories~\citep{wang2023selfinstruct}.
These lines supply the runtime and the raw material for agent training, but they leave open which experience the agent should practice next.
BaT combines both: Stage Bank writes leakage-checked synthetic tasks in the Self-Instruct spirit, executable sandboxes support training, and a fixed OpenHands layer runs the trained policy.

\section{Conclusion}
\label{sec:conclusion}

Benchmark-as-Teacher turns public benchmark stages and checks into an RSI system.
Its asynchronous Stage Bank pipeline builds content-isolated E2E, S-target, and S-mix states, while BiCuRL uses aggregate diagnostics to select a weak stage and update the policy.
The data boundary keeps task-specific evaluation content outside training while aggregate scores guide routing.
\BaTFourBModel{} and \BaTNineBModel{} more than double their corresponding Qwen Instruct Overall scores on AutoMedBench-Lite.
\BaTNineBAgent{} reaches \BaTAgentNineOverall{} Overall and exceeds Claude Opus 4.6 with Claude Code by \BaTOpusMargin{} points.
On external benchmarks, the 9B policy stays within 3.4--5.8 points on AIME and GPQA-Diamond and gains on three of five long-horizon tasks; the 4B policy scores lower on all eight tests.
Together, Stage Bank, BiCuRL, and the fixed BaT Agent execution layer show how a structured benchmark can serve as both evaluator and post-training teacher.

\clearpage
\appendix

\section{Technical Details}
\label{sec:supplementary}

\subsection{Stage Bank Construction}
\label{sec:supp-stagebank}

\paragraph{Row schema.}
Stage Bank is a versioned set $\mathcal{Z}$ of rows.
Each row is a tuple
\begin{equation}
    z=(p,\,s,\,c,\,\mathcal{G},\,\kappa,\,m),
    \label{eq:supp-row}
\end{equation}
where $s\in\mathcal{S}\cup\{\mathrm{E2E}\}$ keys the row to a stage or the end-to-end pool, and $p$ stores the executable sandbox state.
The public benchmark rubric supplies item set $c$, and $\mathcal{G}$ lists the evidence requirements used to compute $\eta_{i,k}$~\citep{liu2026automedbench}.
$\kappa$ stores the attached stage skill, and $m$ records row provenance.
The three pools in Equation~\ref{eq:bicurl-curriculum} are
$\mathcal{Z}_{\mathrm{target}}(s_r^\star)=\{z:s=s_r^\star\}$,
$\mathcal{Z}_{\mathrm{mix}}(s_r^\star)=\{z:s\in\mathcal{S}\setminus\{s_r^\star\}\}$, and
$\mathcal{Z}_{\mathrm{E2E}}=\{z:s=\mathrm{E2E}\}$.

\paragraph{Teacher trajectories and SFT slices.}
For fictional task $\xi_i$, a teacher model produces a checked multi-turn trajectory
\begin{equation}
    \tau_i^{\mathrm{T}}
    =\big((a_{i,t}^{\mathrm{T}},o_{i,t}^{\mathrm{T}},s_{i,t})\big)_{t=1}^{T_i},
    \label{eq:supp-teacher-trajectory}
\end{equation}
where $a_{i,t}^{\mathrm{T}}$ is one teacher response, $o_{i,t}^{\mathrm{T}}$ is the environment observation that follows it, and $s_{i,t}$ is the public workflow stage.
The history $h_{i,t}^{\mathrm{T}}$ contains the task, stage skill, and all earlier responses and observations.
Stage Bank converts the trajectory into single-response rows
\begin{equation}
    \mathcal{D}_{\mathrm{SFT}}
    =\left\{\left(h_{i,t}^{\mathrm{T}},a_{i,t}^{\mathrm{T}}\right):
    1\leq t\leq T_i,\;
    \xi_i\ \text{and}\ \tau_i^{\mathrm{T}}\ \text{pass validation}\right\}.
    \label{eq:supp-sft-data}
\end{equation}
Each row keeps the multi-turn history but applies loss only to the selected teacher response~\citep{xu2025agenttrek,chen2026agenticdpo}:
\begin{equation}
    \mathcal{L}_{\mathrm{SFT}}(\theta)
    =-\mathbb{E}_{(h,a)\sim\mathcal{D}_{\mathrm{SFT}}}
    \left[\sum_{j=1}^{|a|}\log\pi_\theta(a_j\mid h,a_{<j})\right].
    \label{eq:supp-sft-loss}
\end{equation}
Stage Bank excludes tool observations and other environment text from the loss.

\paragraph{Stage sandbox construction.}
Let $b_{i,s}$ denote the turn where stage $s$ begins in $\tau_i^{\mathrm{T}}$.
A stage sandbox for $s$ replays the trajectory prefix inside the execution environment:
\begin{equation}
    p_{i,s}=\Phi_{\xi_i}\big((a_{i,t}^{\mathrm{T}},o_{i,t}^{\mathrm{T}})_{t<b_{i,s}}\big),
    \label{eq:supp-sandbox}
\end{equation}
where $\Phi_{\xi_i}$ executes the prefix and materializes the resulting files, environment, and intermediate artifacts as the row's start state.
Training therefore begins at the stage boundary with an upstream context, while the row rubric scores only the work of stage $s$.

\paragraph{E2E surrogate construction.}
Stage Bank first applies a reduction operator to the fictional task:
\begin{equation}
    \widetilde{\xi}_i
    =R_{\boldsymbol{\lambda}}(\xi_i),\qquad
    \boldsymbol{\lambda}
    =(\lambda_{\mathrm{case}},\lambda_{\mathrm{input}},\lambda_{\mathrm{validate}}).
    \label{eq:supp-e2e-reduction}
\end{equation}
$R_{\boldsymbol{\lambda}}$ retains $\lambda_{\mathrm{case}}\in\{1,\ldots,N_i\}$ of the task's $N_i$ cases, caps each input at a fraction $\lambda_{\mathrm{input}}\in(0,1]$ of its source size, and permits at most $\lambda_{\mathrm{validate}}\in\mathbb{N}_{+}$ validation passes.
The Data Factory chooses $\boldsymbol{\lambda}$ from a versioned task-specific grid and records it in provenance $m_i$.
The reduction keeps the five-stage order, task semantics, and output schema.
The teacher then executes the reduced task and produces a checked trajectory
\begin{equation}
    \widetilde{\tau}_i^{\mathrm{T}}
    =\big((\widetilde{u}_{i,q}^{\mathrm{T}},
    \widetilde{o}_{i,q}^{\mathrm{T}},
    \widetilde{s}_{i,q})\big)_{q=1}^{\widetilde{T}_i},
    \label{eq:supp-e2e-trajectory}
\end{equation}
where $\widetilde{T}_i$ is the number of interaction turns after reduction.
The E2E sandbox starts before the first stage, $p_i^{\mathrm{E2E}}=\Phi_{\widetilde{\xi}_i}(\emptyset)$, and attaches the ordered skills $(\kappa_{i,s})_{s\in\mathcal{S}}$.
Its rubric and evidence contracts combine the stage contracts:
\begin{equation}
    c_i^{\mathrm{E2E}}=\bigcup_{s\in\mathcal{S}}c_{i,s},
    \qquad
    \mathcal{G}_i^{\mathrm{E2E}}=\bigcup_{s\in\mathcal{S}}\mathcal{G}_{i,s}.
    \label{eq:supp-e2e-contract}
\end{equation}
Stage Bank accepts the surrogate only when the sandbox starts, the checked teacher trajectory fits the rollout limit $H$, every stage contract passes, and the leakage scan returns zero:
\begin{equation}
    \begin{aligned}
    \operatorname{Accept}(z_i^{\mathrm{E2E}})
    ={}&\mathbb{I}[\operatorname{Boot}(p_i^{\mathrm{E2E}})=1]
    \mathbb{I}[\widetilde{T}_i\leq H]\\
    &{}\times\prod_{s\in\mathcal{S}}
    \mathbb{I}[\operatorname{Check}_s(\widetilde{\tau}_i^{\mathrm{T}})=1]\\
    &{}\times\mathbb{I}[\operatorname{Leak}(z_i^{\mathrm{E2E}})=0].
    \end{aligned}
    \label{eq:supp-e2e-accept}
\end{equation}
The accepted row preserves the full workflow while keeping each rollout small enough for repeated training.

\subsection{Multi-Turn Supervision in BiCuRL}
\label{sec:supp-multiturn}

Each rollout $y_{i,k}$ is a multi-turn interaction.
At turn $q$, the policy emits response $u_{i,k,q}$ conditioned on the sandbox state, stage skill, and interaction history, after which the sandbox returns observation $o_{i,k,q}$.
The rubric verifier scores the completed rollout, so one rollout-level reward must reach every token that produced it.
For $Q_{i,k}$ interaction turns, the policy and environment generate
\begin{equation}
    P_\theta(y_{i,k},o_{i,k}\mid z_i)
    =\prod_{q=1}^{Q_{i,k}}
    \pi_\theta(u_{i,k,q}\mid h_{i,k,q})
    P_{\mathcal{E}}(o_{i,k,q}\mid h_{i,k,q},u_{i,k,q}).
    \label{eq:supp-multiturn-rollout}
\end{equation}
After tokenizing and joining the $Q_{i,k}$ policy responses, we write $y_{i,k}=(a_{i,k,1},\ldots,a_{i,k,T_{i,k}})$ for the generated tokens and exclude observation tokens from the loss.
Let $h^{\mathrm{tok}}_{i,k,t}$ be the full token context before $a_{i,k,t}$; it contains $p_i$, $\kappa_i$, earlier generated tokens, and every observation returned before that token.

For each generated token, the importance ratio compares the updated policy with the policy that produced the rollout:
\begin{equation}
    w_{i,k,t}
    =\frac{\pi_{\theta}(a_{i,k,t}\mid h^{\mathrm{tok}}_{i,k,t})}
    {\pi_{\bar{\theta}_r}(a_{i,k,t}\mid h^{\mathrm{tok}}_{i,k,t})}.
    \label{eq:supp-ratio}
\end{equation}
Each token receives rollout advantage $A_{i,k}$ from Equation~\ref{eq:bicurl-reward}, clipped within $\epsilon$:
\begin{equation}
    g_{i,k,t}
    =\min\!\Big(w_{i,k,t}A_{i,k},\;
    \operatorname{clip}\big(w_{i,k,t},1\!-\!\epsilon,1\!+\!\epsilon\big)A_{i,k}\Big).
    \label{eq:supp-clip}
\end{equation}
The inner objective of Equation~\ref{eq:bicurl-bilevel} averages these token terms over each rollout and group, with a KL penalty toward the current policy~\citep{shao2024deepseekmath}:
\begin{equation}
\begin{split}
\mathcal{J}_{\mathrm{GRPO}}(\theta;\mathbf{q}_r)
&=\mathbb{E}_{z\sim\mathbf{q}_r}
\Bigg[\frac{1}{K}\sum_{k=1}^{K}\frac{1}{T_{i,k}}
\sum_{t=1}^{T_{i,k}}g_{i,k,t}\Bigg]\\
&\quad-\beta D_{\mathrm{KL}}\!\left(\pi_{\theta}\,\|\,\pi_{\bar{\theta}_r}\right).
\label{eq:supp-grpo}
\end{split}
\end{equation}
Each generated token of rollout $y_{i,k}$ receives $A_{i,k}$, so one group-normalized and evidence-discounted rubric decision supervises every turn.
The curriculum $\mathbf{q}_r$ decides which stage those trajectories practice.

\subsection{Migration to Other Staged Benchmarks}
\label{sec:supp-migration}

BiCuRL uses the staged benchmark $\mathcal{B}=(\mathcal{S},\mathcal{C},\mathcal{V})$ defined in Section~\ref{sec:method-data}.
A compatible benchmark supplies ordered stage boundaries in $\mathcal{S}$, public rubric and evidence contracts in $\mathcal{C}$, and fixed evaluation $\mathcal{V}(\pi)=(M,\mathbf{e})$.

\begin{proposition}
Suppose Stage Bank can materialize executable states for every stage in $\mathcal{S}$.
If policy $\pi_\theta$ exposes the token likelihoods required by GRPO and policy KL, the BiCuRL update is well-defined for $\mathcal{B}$.
\end{proposition}

\emph{Argument.}
The router reads only $\mathbf{e}_r$ and selects $s_r^\star\in\mathcal{S}$.
Equation~\ref{eq:bicurl-curriculum} partitions Stage Bank by stage key, with S-mix drawing the remaining stages.
Stage Bank construction uses the assumed stage boundaries, executable states, and public rubric contracts.
Equations~\ref{eq:bicurl-reward} and~\ref{eq:supp-grpo} use rubric items and evidence requirements regardless of the number or meaning of stages.
Equation~\ref{eq:bicurl-update} uses the scalar score sequence and the assumed policy KL.
The likelihood assumption also defines the GRPO ratios in Equation~\ref{eq:supp-ratio}.
These inputs define every term in the update. \hfill$\square$

This proposition establishes procedural compatibility.
Whether a migrated BaT system improves a particular benchmark remains an empirical question, consistent with the bounded transfer results in the main text.

\subsection{Post-Training and Runtime Details}
\label{sec:supp-training-details}

\paragraph{Cold-start SFT.}
We train the Qwen3.5-9B cold-start policy with full-parameter supervised fine-tuning on \BaTNineBSFTRows{} SFT rows~\citep{qwen2026qwen359b,ouyang2022training}.
Training uses eight GPUs with PyTorch Fully Sharded Data Parallel (FSDP) \texttt{full\_shard} and fused AdamW~\citep{zhao2023fsdp,loshchilov2019adamw}.
Table~\ref{tab:sft-config} reports the key configuration.
BiCuRL uses the final archived SFT checkpoint as its initializer.

\begin{table}[!htbp]
    \centering
    {\scriptsize
    \setlength{\tabcolsep}{3.0pt}
    \renewcommand{\arraystretch}{1.06}
    \begin{tabular}{@{}ll@{}}
        \toprule
        \textbf{Parameter} & \textbf{Qwen3.5-9B cold SFT} \\
        \midrule
        Parameter update & Full parameters, without adapters \\
        Parallelism & 8 GPUs, FSDP \texttt{full\_shard} \\
        Per-GPU microbatch & 1 \\
        Gradient accumulation & 8 \\
        Effective global batch & 64 \\
        Learning rate & $2\times10^{-5}$ \\
        Schedule / warmup & Cosine / 3\% \\
        Optimizer & Fused AdamW \\
        Weight decay / max grad norm & 0 / 1 \\
        Target training budget & 200 steps \\
        Checkpoint interval & 5 steps \\
        SFT rows / evaluation & \BaTNineBSFTRows{} / disabled \\
        Seed & 42 \\
        \bottomrule
    \end{tabular}
    }
    \caption{
        \textbf{Qwen3.5-9B cold-start SFT configuration.}
        We use full-parameter training with PyTorch FSDP and AdamW~\citep{qwen2026qwen359b,zhao2023fsdp,loshchilov2019adamw}.
    }
    \label{tab:sft-config}
\end{table}

\paragraph{Pool weights and sampling.}
Each round draws rows from the three Stage Bank pools with proportions
$(\rho_{\mathrm{target}},\rho_{\mathrm{mix}},\rho_{\mathrm{E2E}})=(1/2,1/4,1/4)$.
Within E2E, Stage Bank draws surrogate rows uniformly.
Sampling is without replacement inside one round and resets between rounds.
Matched ablation runs draw from the \BaTMatchedAblationRows{}-row pool described in Section~\ref{sec:setup} and renormalize proportions after omitting a pool.

\paragraph{Fallback KL threshold.}
The policy-shift trigger in Equation~\ref{eq:bicurl-update} uses threshold $\tau=0.1$.
We estimate it as the mean per-token KL divergence between candidate policy $\pi_{\widetilde{\theta}_{r+1}}$ and retained policy $\pi_{\theta^\star}$ over 256 Stage Bank prompts held out from training.
The patience trigger uses three consecutive score drops.

\paragraph{Serving stack and throughput.}
We run training and agent rollouts on eight NVIDIA A100 GPUs with 80GB memory per GPU~\citep{nvidia2022a100}.
The serving layer supports SGLang and vLLM~\citep{zheng2024sglang,kwon2023vllm}.
We use SGLang by default for agentic tasks and set the serving context limit to 256K tokens.
Matched ablations cap each training sequence at 12,288 tokens, as Table~\ref{tab:ablation-config} records.
The source run summary records one node-level throughput value per model on the eight-GPU node: 320 tokens/s for Qwen3.5-4B and 220 tokens/s for Qwen3.5-9B.
The summary combines prefill and decoding and omits the remaining serving controls, so we treat both rates as descriptive measurements.

\begin{table}[!htbp]
    \centering
    {\scriptsize
    \setlength{\tabcolsep}{3.0pt}
    \renewcommand{\arraystretch}{1.06}
    \begin{tabular}{@{}lcc@{}}
        \toprule
        \textbf{Runtime setting} & \textbf{Qwen3.5-4B} & \textbf{Qwen3.5-9B} \\
        \midrule
        GPUs & \multicolumn{2}{c}{8 $\times$ NVIDIA A100 80GB} \\
        Default engine & \multicolumn{2}{c}{SGLang} \\
        Supported engine & \multicolumn{2}{c}{vLLM} \\
        Context limit & \multicolumn{2}{c}{256K tokens} \\
        Recorded throughput & 320 tokens/s & 220 tokens/s \\
        \bottomrule
    \end{tabular}
    }
    \caption{
        \textbf{Serving configuration and recorded throughput.}
        The stack supports SGLang and vLLM, with SGLang as the default engine for agentic tasks~\citep{zheng2024sglang,kwon2023vllm}.
        Each recorded rate uses an eight-GPU NVIDIA A100 80GB node~\citep{nvidia2022a100}.
        The source run summary reports an unsplit token rate and omits the remaining serving controls.
    }
    \label{tab:serving-config}
\end{table}

\subsection{Local LLM Comparison}
\label{sec:supp-local-llms}
Local deployment helps keep medical data within the research environment.
We define local LLMs as models with fewer than 40B parameters that can run on a single NVIDIA A100.
We downloaded and deployed 19 representative open-weight models, then evaluated them with the same default runner and benchmark protocol~\citep{wang2025openhands}.
The default runner alone produces these results; the BaT agent harness plays no role.
\tableautorefname~\ref{tab:local-llm-results} shows that \BaTNineBModel{} achieves 53.4 Overall and ranks third.
The two higher-ranked models have 27B and 35B total parameters, which suggests that model scale contributes to the remaining gap.
Future work will study how BaT scales to larger LLM backbones.

\begin{table}[!htbp]
    \centering
    {\scriptsize
    \setlength{\tabcolsep}{5pt}
    \renewcommand{\arraystretch}{1.06}
    \begin{tabular}{@{}lrrr@{}}
        \toprule
        \textbf{Model} & \textbf{Overall} & \textbf{Agentic} & \textbf{Task} \\
        \midrule
        Qwen3.5-4B~\citep{qwen2026qwen354b} & 6.1 & 12.1 & 0.0 \\
        Meissa-4B~\citep{chen2026meissa} & 7.6 & 15.2 & 0.0 \\
        Gemma 4 E2B~\citep{gemmateam2026gemma4} & 9.0 & 18.0 & 0.0 \\
        Nemotron 3 Nano 4B~\citep{nvidia2026nemotron3nano4b} & 9.1 & 18.2 & 0.0 \\
        GPT-OSS-20B~\citep{openai2025gptoss} & 17.6 & 25.3 & 9.9 \\
        Nemotron 3 Nano 30B-A3B~\citep{nvidia2025nemotron3nano} & 18.0 & 22.5 & 13.5 \\
        Nemotron Nano 9B v2~\citep{nvidia2025nemotronnano9bv2} & 18.7 & 24.6 & 12.8 \\
        Qwen3.5-9B~\citep{qwen2026qwen359b} & 19.9 & 11.3 & 28.4 \\
        Gemma 4 E4B~\citep{gemmateam2026gemma4} & 21.9 & 24.2 & 19.6 \\
        \rowcolor{batrowgreen}
        BaT-4B & 22.9 & 28.8 & 17.0 \\
        Gemma 4 12B~\citep{gemmateam2026gemma4} & 23.1 & 20.2 & 26.0 \\
        ClinSeek-35B~\citep{wu2026clinseek} & 26.0 & 39.7 & 12.3 \\
        Gemma 4 31B~\citep{gemmateam2026gemma4} & 30.4 & 29.0 & 31.8 \\
        Gemma 4 26B-A4B~\citep{gemmateam2026gemma4} & 33.9 & 33.8 & 34.0 \\
        Muse-Glimmer-30B~\citep{meta2026museglimmer} & 36.5 & 44.0 & 29.0 \\
        Nemotron 3.5 Lightning 30B-A3B~\citep{nvidia2026nemotron35lightning} & 39.7 & 47.0 & 32.4 \\
        \rowcolor{batrowgreen}
        BaT-9B & 53.4 & 64.7 & 42.1 \\
        Qwen3.5-35B-A3B~\citep{qwen2026qwen3535ba3b} & 57.9 & 68.8 & 47.0 \\
        Qwen3.6-27B~\citep{qwen2026qwen3627b} & 65.3 & 73.3 & 57.3 \\
        \bottomrule
    \end{tabular}
    }
    \caption{
        \textbf{Full local LLM comparison with the default runner.}
        Scores use the AutoMedBench-Lite evaluation~\citep{liu2026automedbench} and a 0--100 scale, ranked by Overall score.
        The two shaded rows are BaT policies.
        Overall equals the arithmetic mean of Agentic and Task scores.
    }
    \label{tab:local-llm-results}
\end{table}

\subsection{AutoMedBench-Lite Protocol}
\label{sec:supp-protocol}

AutoMedBench-Lite applies the five-stage workflow to seven held-out medical AI tracks~\citep{liu2026automedbench}.
Table~\ref{tab:automedbench-lite-protocol} distinguishes each track's public case pool from the ten separate agent executions used to evaluate a system on that track.

\begin{table}[!htbp]
    \centering
    {\scriptsize
    \setlength{\tabcolsep}{2.2pt}
    \renewcommand{\arraystretch}{1.06}
    \begin{tabular}{@{}llrr@{}}
        \toprule
        \textbf{Track} & \textbf{Held-out task} & \textbf{Cases} & \textbf{Runs} \\
        \midrule
        Classification & Skin-lesion classification & 100 & 10 \\
        Detection & Wrist radiograph detection & 100 & 10 \\
        Enhancement & Low-dose CT denoising & 20 & 10 \\
        Report & Chest X-ray reporting & 100 & 10 \\
        Segmentation & Multi-organ CT segmentation & 40 & 10 \\
        Synthesis & Pancreas CT synthesis & 20 & 10 \\
        VQA & Multimodal medical VQA & 2,005 & 10 \\
        \bottomrule
    \end{tabular}
    }
    \caption{
        \textbf{AutoMedBench-Lite spans seven held-out medical AI tracks and 70 runs per system.}
        Cases report the public evaluation subsets, and each system runs every track ten times.
        Agentic weights Plan, Setup, Validate, Inference, and Submit by 0.25/0.15/0.35/0.15/0.10; Overall averages Task and Agentic~\citep{liu2026automedbench}.
    }
    \label{tab:automedbench-lite-protocol}
\end{table}

Each run yields a Task score for its submitted result and stage scores for Plan, Setup, Validate, Inference, and Submit.
Agentic combines the five stage scores with the weights in the table caption, and Overall gives Task and Agentic equal weight.

\paragraph{Aggregation and uncertainty.}
The statistical unit is the track.
For each system, we first average each metric over the ten repeats of a track and then report the mean of the seven track scores.
We treat the seven track means as the independent units because runs within a track share task content.
We pair comparisons between two systems at the track level.
The controller evaluation during training and the final evaluation share these seven tracks but use disjoint runs; controller-round scores never enter the reported tables.

\begin{figure}[!htbp]
    \centering
    \includegraphics[width=\columnwidth]{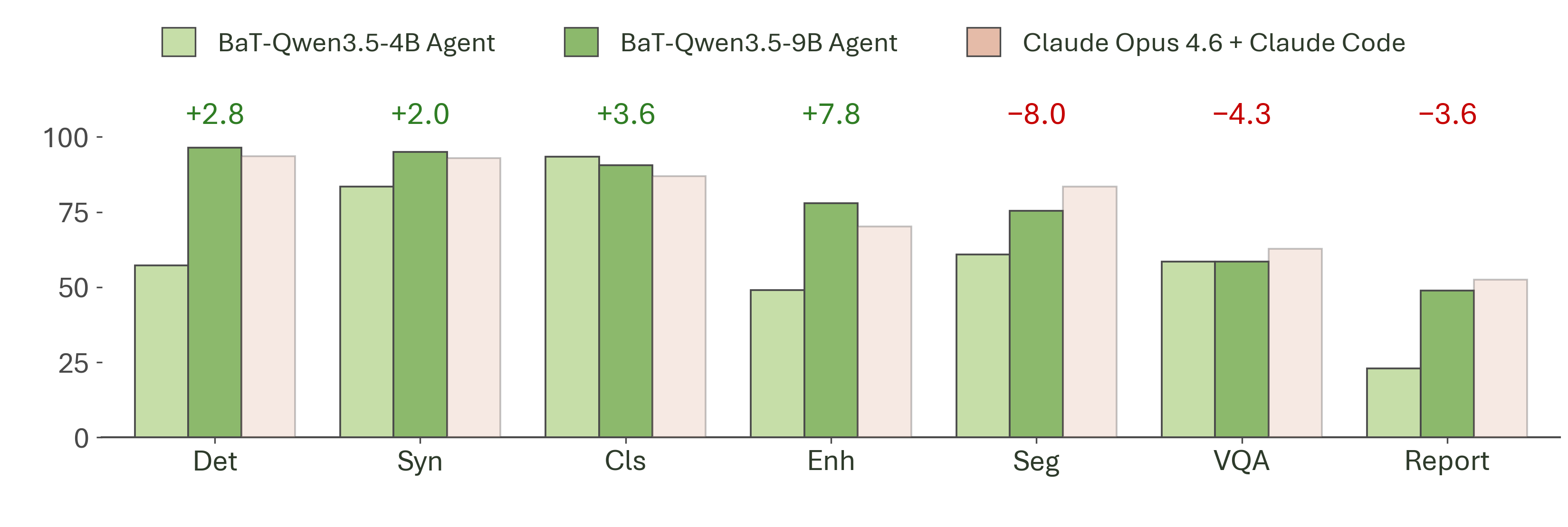}
    \caption{
        \textbf{Per-track Overall comparison of the BaT Agents and Claude Opus 4.6 with Claude Code on AutoMedBench-Lite.}
        Bars show the mean Overall score per track on a 0--100 scale.
        We order tracks left to right by ascending average output tokens per run, from the lightest question-answering tracks to the heaviest imaging pipelines.
        We estimate the Opus enhancement and report values from its published aggregate scores and measure the remaining values.
    }
    \label{fig:downstream-score-tracks}
\end{figure}

\subsection{Matched Ablation Contract}
\label{sec:supp-ablation}

The matched pool-ablation contract changes one variable: which of S-target, S-mix, and E2E enters GRPO.
Table~\ref{tab:ablation-config} records the controls for the \BaTMatchedAblationRows{}-row matched pool.

\begin{table}[!htbp]
    \centering
    {\scriptsize
    \setlength{\tabcolsep}{3.0pt}
    \renewcommand{\arraystretch}{1.06}
    \begin{tabular}{@{}ll@{}}
        \toprule
        \textbf{Parameter} & \textbf{Recorded setting} \\
        \midrule
        Initializer & Qwen3.5-9B SFT checkpoint\\
        Train minibatch & 8 / 8 \\
        Rollouts per prompt & 4 \\
        Actor learning rate & $1\times10^{-6}$ \\
        Actor KL coefficient & $1\times10^{-3}$ \\
        Rollout temperature & 0.7 \\
        Prompt / response limit & 8,192 / 2,048 tokens \\
        Model context / max sequences & 12,288 / 16 \\
        Training seed & 42 \\
        Optimizer between rounds & Reset \\
        Adaptive schedule & 10 rounds $\times$ 50 steps \\
        Final evaluation & 7 tracks $\times$ 10 repeats \\
        \bottomrule
    \end{tabular}
    }
    \caption{Recorded controls for the matched Qwen3.5-9B pool-ablation contract.}
    \label{tab:ablation-config}
\end{table}

\subsection{External Benchmark Protocol}
\label{sec:supp-external}

We evaluate the Instruct baselines and BiCuRL policies on eight external benchmarks.
AIME 2025 and AIME 2026 test competition-math reasoning with integer-answer problems~\citep{maa2026aime}.
GPQA-Diamond tests graduate-level biology, physics, and chemistry questions~\citep{rein2023gpqa}.
$\tau^{2}$-Bench tests multi-turn agents in environments where the agent and user can both take actions~\citep{barres2025tau2bench}.
BFCL-Parity tests function calling, GAIA tests general assistants, SWE-bench Verified tests software issue resolution, and Terminal Bench 2.0 tests command-line work~\citep{patil2025bfcl,mialon2024gaia,openai2024swebenchverified,merrill2026terminalbench}.
Within each row, Base and BiCuRL use the same evaluation setting.
We compute deltas as BiCuRL minus Base in percentage points.

\bibliographystyle{plainnat}
\bibliography{paper}

\end{document}